\documentclass[final,1p,times]{elsarticle}

\usepackage{amssymb}
\usepackage{amsmath}
\usepackage{xcolor}
\usepackage{graphicx}
\usepackage{booktabs}
\usepackage{caption}
\usepackage{arydshln}
\usepackage{multirow}
\usepackage{kotex}
\usepackage{url}
\usepackage{hyperref}

\journal{Automation in Construction}

\begin{document}

\begin{frontmatter}

\title{SlumpGuard: An AI-Powered Real-Time System for Automated Concrete Slump Prediction via Video Analysis}

\author{Youngmin Kim${}$~\fnref{label1}}
\author{Giyeong Oh~\fnref{label1}}
\author{Kwangsoo Youm${}^\dagger$~\fnref{label2}}
\author{Youngjae Yu${}^\dagger$~\fnref{label3}}


\affiliation[label1]{organization={Yonsei University},
            addressline={50 Yonsei-ro, Seodaemun-gu}, 
            city={Seoul},
            postcode={03772}, 
            country={Republic of Korea}}

\affiliation[label2]{organization={GS Engineering \& Construction Corp},
            addressline={33 Jong-ro, Jongno-gu},
            city={Seoul},
            postcode={03159},
            country={Republic of Korea}}

\affiliation[label3]{organization={Seoul National University},
            addressline={1, Gwanak-ro, Gwanak-gu},
            city={Seoul},
            postcode={08826},
            country={Republic of Korea}}

\begin{abstract}
Concrete workability is essential for construction quality, with the slump test being the most widely used on-site method for its assessment. However, traditional slump testing is manual, time-consuming, and highly operator-dependent, making it unsuitable for continuous or real-time monitoring during placement. To address these limitations, we present \textsc{SlumpGuard}, an AI-powered vision system that analyzes the natural discharge flow from a mixer-truck chute using a single fixed camera. The system performs automatic chute detection, pouring-event identification, and video-based slump classification, enabling quality monitoring without sensors, hardware installation, or manual intervention. We introduce the system design, construct a site-replicated dataset of over 6,000 video clips, and report extensive evaluations demonstrating reliable chute localization, accurate pouring detection, and robust slump prediction under diverse field conditions. An expert study further reveals significant disagreement in human visual estimates, highlighting the need for automated assessment. Demonstration videos are available at \href{https://winston1214.github.io/SlumpGuard}{this URL}.
\end{abstract}





\end{frontmatter}



\section{Introduction}
\label{sec:introduction}
\def\thefootnote{$\dagger$}\footnotetext{Co-Supervision Author}
\def\thefootnote{}\footnotetext{\textit{Email address:} \texttt{winston1214@yonsei.ac.kr} (Youngmin Kim)}
Concrete stands as the backbone of modern construction infrastructure, with over 14 billion cubic meters produced annually worldwide, making it the most widely consumed material after water~\cite{hoang2016:estimating}. Among the critical properties governing concrete performance, workability, which is primarily assessed through slump testing, represents a fundamental parameter that directly influences mixing efficiency, transportation logistics, placement operations, and compaction effectiveness on construction sites~\cite{chen2022:research, ojala2024:estimating}. However, the complex and dynamic nature of construction environments presents unprecedented challenges for maintaining consistent concrete quality control. In particular, conventional manual slump testing, still the industry standard, often fails to meet the demanding requirements of large-scale, time-sensitive projects.

The construction industry therefore continues to struggle with real-time quality assurance. On-site inspection of fresh concrete is typically performed only intermittently rather than for every delivered batch. For example, the Korean Construction Specification KCS 14 20 10 mandates slump inspection only once per day or once every 120 cubic meters (approximately 20 truckloads), making continuous, batch-level verification infeasible in practice. As a result, quality deviations that arise during mixing, transportation, or placement may remain undetected until concrete is already being poured, leading to delays, increases in cost, and inconsistency due to human measurement variability~\cite{tuan2021:situ}.

To address these challenges, researchers have increasingly explored automated and data-driven methods for assessing fresh concrete properties. As part of these efforts, sensor-based approaches employ piezoelectric, depth, and fiber-optic measurements to quantify workability-related characteristics. Leveraging these sensing modalities, recent work has explored combining the captured signals with deep learning and IoT-based frameworks to enable automated analysis of fresh concrete properties~\cite{han2025:field, kim2018:visualization, govindaraju2025:real}. However, sensor-based solutions heavily rely on hardware installation and calibration, and their practicality is often constrained by on-site regulatory restrictions and variations in equipment conditions across different construction environments.

To overcome these limitations, visual information–based methods for slump estimation have also been proposed. Yucelai \textit{et al.}~\cite{yucelai:ai} introduced a stereo-vision system for predicting slump during laboratory slump tests. Other studies attempted to estimate slump by installing cameras in proximity to the concrete mixer and analyzing captured images~\cite{idrees2024:automatic, hao2025:concrete}. Despite their promise, these approaches face significant barriers to real-world deployment. The method of Yucelai \textit{et al.}~\cite{yucelai:ai} requires repeated on-site slump testing, which is rarely performed during routine concrete placement, limiting its practical applicability. Methods based on mixer-mounted cameras~\cite{idrees2024:automatic, hao2025:concrete} are also susceptible to dust and slurry contamination, leading to frequent maintenance and reduced operational reliability. Moreover, image-only frameworks generally fail to capture the flow-driven characteristics of fresh concrete, making it difficult to reflect the inherent dynamic behavior underlying slump.


To address the limitations of previous automated approaches, we introduce \textsc{SlumpGuard}, a fully autonomous, camera-only vision system designed for real-site deployment. Operating with a single fixed camera that analyzes the natural discharge flow, \textsc{SlumpGuard} avoids hardware installation and contamination issues and integrates seamlessly into existing workflows. This lightweight, non-intrusive design enables continuous, batch-level monitoring under diverse on-site conditions. At its core, \textsc{SlumpGuard} is built upon a modular three-stage pipeline that structures the overall task into interpretable components.

To operationalize this goal in real construction environments, \textsc{SlumpGuard} decomposes the problem into three stages, each addressing a distinct challenge encountered during concrete placement.
First, the system performs robust chute detection to automatically localize the discharge chute in unconstrained on-site videos. This step isolates the region where concrete flows, enabling stable analysis regardless of camera placement, background clutter, or chute rotation.
Second, \textsc{SlumpGuard} identifies both when and from which chute concrete begins to fall by analyzing pixel-level motion using optical flow. This automated detection of flow onset ensures that the system processes only frames containing concrete discharge, eliminating dependence on manual trimming or operator input.
Third, once the relevant video segment is identified, a video-based slump classification module analyzes the temporal evolution of concrete flow to estimate the slump category in real time. By repeatedly predicting slump over successive frame windows and aggregating the results, the system achieves stable and reliable batch-level slump assessment. This modular design enhances interpretability and facilitates adaptation to different site conditions, while the motion-gated processing ensures real-time operation by restricting inference to concrete–flow intervals.

To assess the practical effectiveness of this three-stage pipeline and the deployability of \textsc{SlumpGuard}, we conducted extensive experiments on more than 6,000 video clips collected under site-replicated conditions.
Our evaluation shows that each module contributes complementary strengths toward stable real-world operation.
The chute detection module (Stage 1) consistently identified mixer-truck chutes across diverse geometric and illumination conditions, while also achieving high reliability in recognizing which chute was actively discharging concrete (Stage 2). The video-based slump classification module (Stage 3) further demonstrated strong performance, achieving over 80\% accuracy despite substantial visual variability in concrete flow. This variability underscores the practical need for objective, automated systems capable of consistent assessment in conditions where human judgment is inherently limited. Together, these findings indicate that \textsc{SlumpGuard} offers a reliable and field-ready vision-based alternative to conventional manual or sensor-dependent inspection methods.

The remainder of this paper is organized as follows. Section~\ref{sec:relatedwork} reviews existing studies on concrete workability assessment and automated monitoring approaches. Section~\ref{sec:dataset} describes the dataset collection process and annotation methodology. Section~\ref{sec:method} details the architecture of the proposed \textsc{SlumpGuard} system, including its three-stage pipeline and data augmentation strategies. Section~\ref{sec:experiments} presents experimental results and performance evaluation under real-site conditions. Section~\ref{sec:deployment} introduces how \textsc{SlumpGuard} is deployed and utilized in real construction sites. Section~\ref{sec:discussion} and Section~\ref{sec:limitation} provide a discussion of \textsc{SlumpGuard}, including its implications and limitations. Finally, Section~\ref{sec:conclusion} concludes the paper and outlines future research directions.

\section{Background and Related Studies}
\label{sec:relatedwork}
\subsection{Optical Flow}
Optical flow is a computer vision technique used to estimate the motion of objects or scenes across video frames by analyzing the movement of pixels. It represents motion as a vector field, where each vector indicates the displacement of a pixel between consecutive frames. There are two main approaches: dense optical flow, which computes motion for every pixel, and sparse optical flow, which estimates motion only at selected feature points. While dense optical flow~\cite{horn1981:determining} offers a detailed view of the entire scene's motion, it is computationally intensive. Sparse optical flow is more efficient but may overlook fine-grained motion.

In this work, we adopt the Lucas-Kanade method~\cite{lucas1981:iterative}, a widely used algorithm for sparse optical flow, suitable for real-time applications such as tracking concrete flow. It assumes that the intensity of a point remains constant over time:
\begin{equation}
I(x,y,t) = I(x + \Delta x, y + \Delta y, t + \Delta t),
\end{equation}
and linearizes this relation using a Taylor series expansion, yielding:
\begin{equation}
\frac{\partial I}{\partial x}u + \frac{\partial I}{\partial y}v + \frac{\partial I}{\partial t} = 0,
\end{equation}
where $u$ and $v$ are the horizontal and vertical components of the flow. By solving this equation over a small window around each feature point, the Lucas-Kanade method efficiently estimates the local motion.

However, classical optical flow methods, including Lucas-Kanade, often struggle under challenging conditions such as illumination changes, occlusions, or non-rigid motion. To overcome these limitations, deep learning-based methods have emerged as the new state-of-the-art.

FlowNet~\cite{dosovitskiy2015:flownet} pioneered the use of convolutional neural networks for end-to-end optical flow prediction, learning motion patterns directly from synthetic datasets. Building on this idea, FlowNet2~\cite{ilg2017:flownet}, PWC-Net~\cite{sun2018:pwc}, and LiteFlowNet~\cite{hui2018:liteflownet} introduced architectural improvements such as pyramid-based warping, cost volume reasoning, and hierarchical refinement. More recently, RAFT~\cite{teed2020:raft} achieved significant performance gains by leveraging dense, all-pairs correlations and iterative updates to produce highly accurate flow estimates.

Previous studies have applied optical flow combined with imaging techniques for slump prediction under controlled experimental settings~\cite{tuan2021:situ, kim2018:visualization}. In contrast, our approach is specifically designed for robust, real-world deployment on construction sites, enabling continuous, automated slump quality monitoring under diverse and dynamic field conditions.

\subsection{Object Detection}
Object detection is a fundamental computer vision task widely applied in construction to automatically identify and localize key entities such as workers, equipment, and materials in images or videos. 
Recent studies have demonstrated its effectiveness in construction site monitoring~\cite{pfitzner2025:monitoring, kim2023:small}.

State-of-the-art detection frameworks such as Faster R-CNN~\cite{ren2015:faster}, Mask R-CNN~\cite{he2017:mask}, SSD~\cite{liu2016:ssd}, and YOLO~\cite{redmon2016:you} offer a good balance between speed and accuracy, making them well-suited for real-time applications on construction sites.

However, in construction environments where objects like materials and equipment often appear at arbitrary angles, standard axis-aligned bounding boxes may be insufficient. In such cases, oriented object detection becomes essential. Early approaches like R2CNN~\cite{jiang2017:r2cnn} introduced rotation-invariant proposals, while more recent models such as Oriented R-CNN~\cite{xie2021:oriented} and ReDet~\cite{han2021:redet} employ rotation-aware architectures to improve detection accuracy. These methods are particularly effective for automated inspection tasks involving rotated or non-axis-aligned objects.

\subsection{Video Classification}
Video classification is a fundamental task in computer vision that involves assigning semantic labels to video clips by analyzing both spatial and temporal information inherent in the data. Recent advances in this domain have been largely driven by two primary classes of deep learning architectures: 3D convolutional neural networks (3D CNNs) and vision transformers~\cite{dosovitskiy2020:image} tailored for video.

3D CNNs like ResNet3D~\cite{tran2018:closer} extend 2D convolutions to the temporal dimension, effectively capturing spatiotemporal features and achieving strong results on benchmarks such as Kinetics-400~\cite{kay2017:kinetics}. Transformer-based models, notably TimeSformer~\cite{bertasius2021:space}, use divided spatial and temporal self-attention to model long-range dependencies efficiently, surpassing many convolutional approaches. Further advancements include VideoSwin~\cite{liu2022:video} and MViT~\cite{fan2021:multiscale}, which incorporate hierarchical and multiscale attention mechanisms to improve video representation learning and classification performance.

These video classification models have also been increasingly applied in the construction industry, playing a pivotal role in streamlining quality supervision, activity recognition, and concrete workability assessment.
Previous studies have also explored video-based approaches for material quality monitoring and evaluation~\cite{guo2025:egocentric, ojala2024:estimating, idrees2024:automatic}, alongside behavior classification systems aimed at improving worker safety and productivity~\cite{li2022:action, yang2023:transformer}.

\section{Dataset}
\label{sec:dataset}
\subsection{Data Collection}
\begin{figure*}
    \centering
    \includegraphics[width=\linewidth]{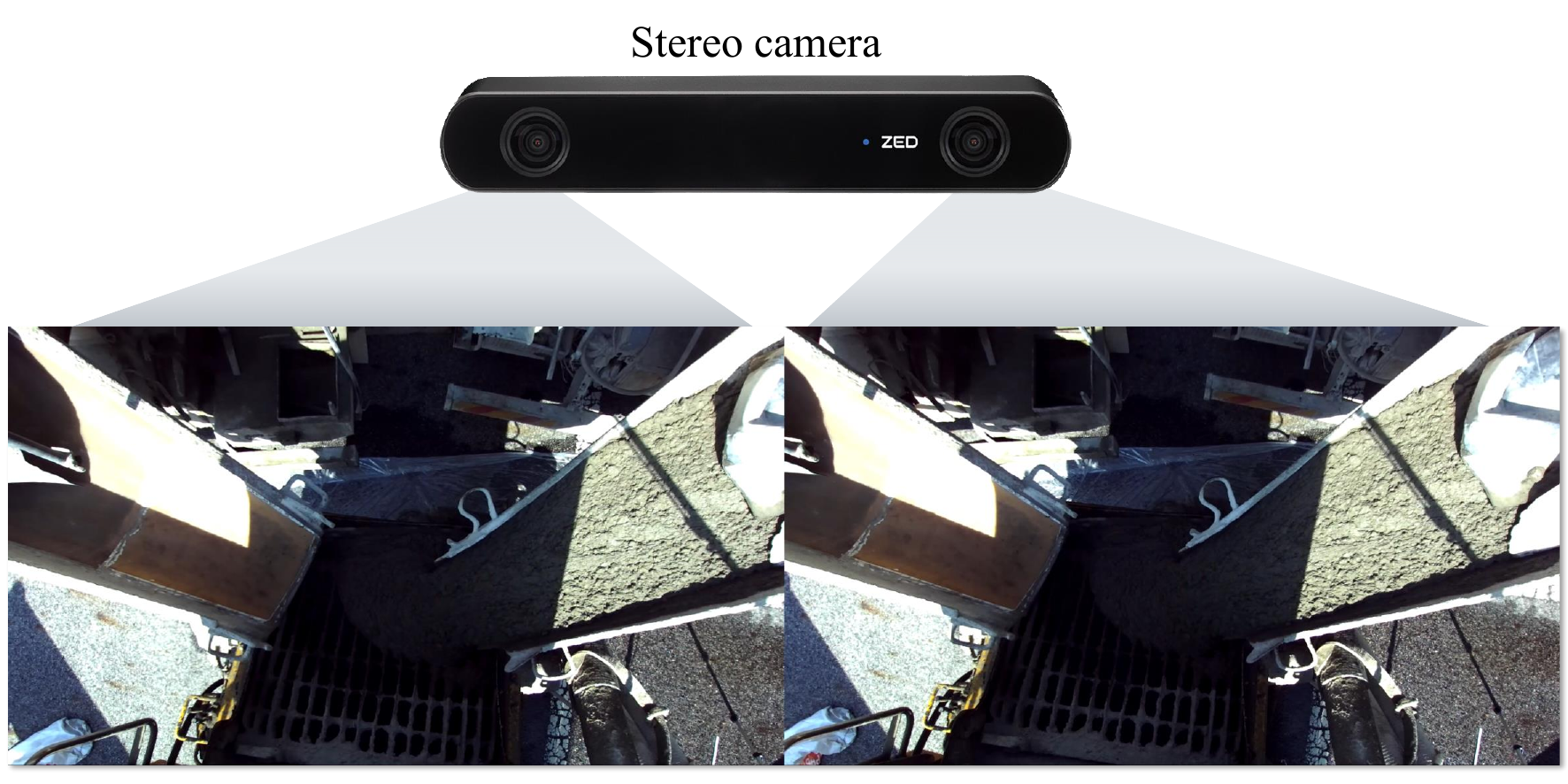}
    \caption{An example of our collected dataset. The concrete pouring from two concrete mixer truck's chutes was captured using a ZED 2i stereo camera. The overall resolution of each camera system is $3840  \times 1080$, consisting of two individual cameras with $1920 \times 1080$ resolution each.}
    \label{fig:example_data}
\end{figure*}

In this study, we constructed a high-quality video dataset to accurately capture the dynamic slump behaviors of ready-mixed concrete during concrete placing. To achieve this, we conducted controlled concrete pouring experiments using two ready-mix concrete trucks, each carrying concrete mixtures with predetermined slump values.  

In the experimental process, considering cost efficiency, we strategically utilized a single stereo camera for video acquisition. The stereo camera allowed simultaneous capturing of video footage from two distinct viewpoints in a single filming session, allowing richer data collection within a limited number of experiments. All videos were recorded at 30 frames per second (FPS), with a stereo capture resolution of $3840 \times 1080$ composed of two individual camera views, each operating at $1920 \times 1080$. This approach maximized data efficiency while clearly capturing the concrete flow through the chute. Additionally, we performed traditional slump cone measurements at regular intervals, establishing precise ground-truth labels corresponding to each video frame.
This example is shown in Figure~\ref{fig:example_data}.

\subsection{Preprocessing and Annotation}
\begin{figure*}[t]
    \centering
    \includegraphics[width=\linewidth]{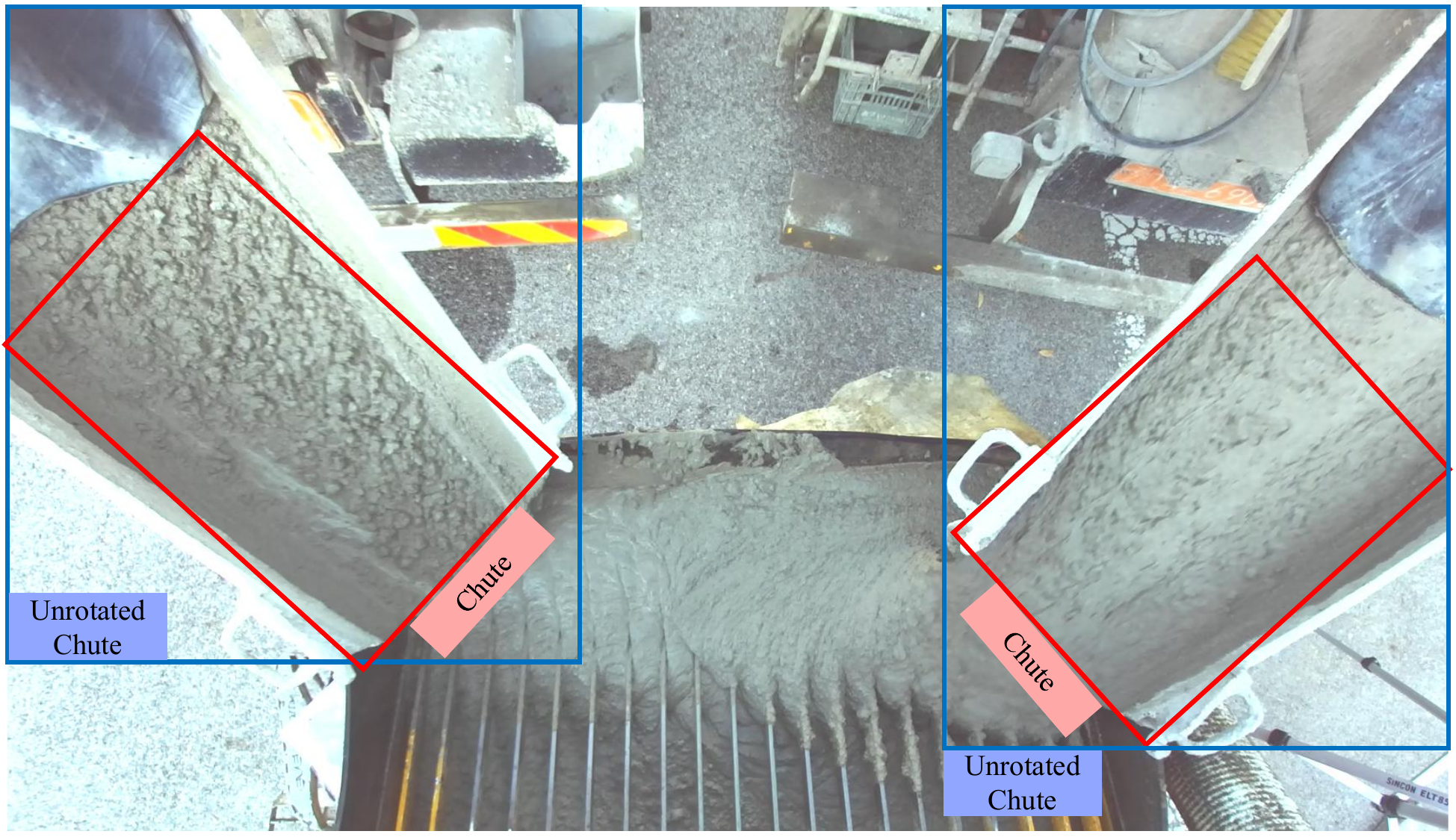}
    \caption{An example of data annotation for detecting concrete pouring regions using bounding boxes. The red boxes represent the chutes, while the blue boxes indicate the corresponding unrotated bounding boxes. }
    \label{fig:ann_data}
\end{figure*}

Since we employed a stereo camera setup, the collected videos consisted of two separate streams. Therefore, as part of preprocessing, the original stereo data was divided into two individual video sequences, effectively doubling the amount of data for analysis. Considering that concrete discharges directly from the ready-mixed concrete chute, our preprocessing specifically targeted the chute region. 
Bounding box annotations were performed by categorizing the chute region into two classes, as shown in Figure~\ref{fig:ann_data}: an ``Unrotated Chute'' and a ``Chute''. The ``Unrotated Chute'' corresponds to an axis-aligned bounding box that fully encloses the chute, ensuring that both endpoints are contained within the box. In contrast, the ``Chute'' class uses rotated bounding boxes to precisely capture the chute’s orientation.

\section{Automation Strategy}
\label{sec:method}
\begin{figure*}[t]
    \centering
    \includegraphics[width=\linewidth]{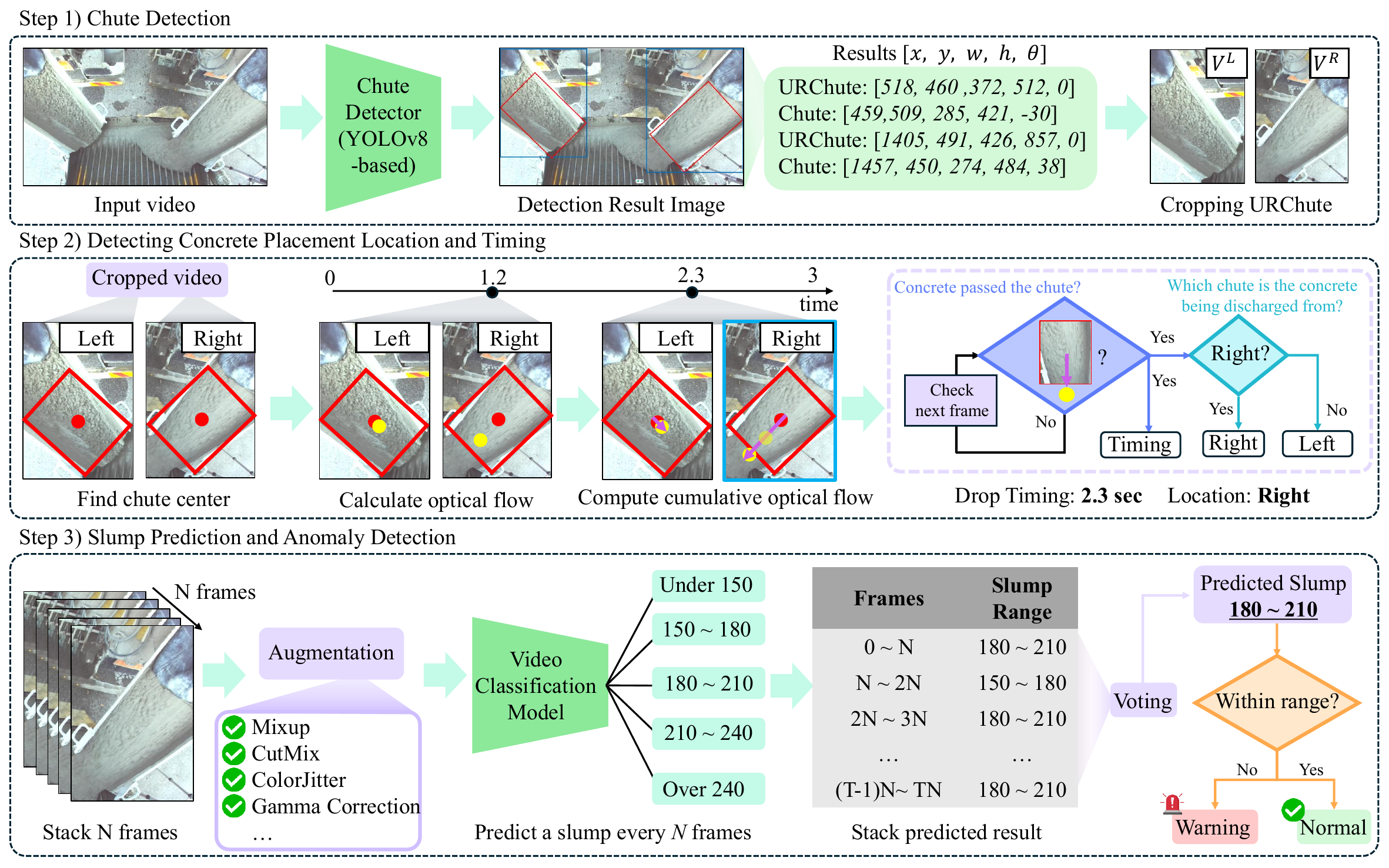}
    \caption{Overview of our pipeline. Our pipeline consists of three steps: detecting chutes, identifying when and from which chute an object falls, and predicting the slump. Finally, we compare the predicted slump with the requested range to perform anomaly detection. In the figure, ``URChute'' refers to an unrotated chute. The variables $x, y, w, h$, and $\theta$ represent the bounding box's center $x$-coordinates, $y$-coordinates, width, and height, respectively, with $\theta$ measured in degrees.}
    \label{fig:pipeline}
\end{figure*}

To practically implement our pipeline at construction sites, we propose automating the entire pipeline. Specifically, the following three aspects will be automated:
\begin{enumerate}
    \item Determining from which of the two chutes the concrete is discharged.
    \item Identify the exact time when the concrete begins to flow from the chute.
    \item Assessing the discharged concrete slump to determine whether it falls within a specified range.
\end{enumerate}
An overview of our proposed automation pipeline is presented in Figure~\ref{fig:pipeline}.

\subsection{Chute Detection} 
\label{subsec:step1}
To automatically identify when concrete is falling from the chute, we first detect the chute region in the video using an object detection model (see Step 1 in Figure~\ref{fig:pipeline}). We employ YOLOv8~\cite{yolov8}, a state-of-the-art object detection model, known for its fast inference performance and high accuracy, making it suitable for real-time object detection. 
We initialize the model with weights pretrained on the DOTA dataset~\cite{xia2018:dota}, which is designed for detecting oriented objects in aerial imagery and is well-suited for our task of detecting rotated chutes.
Using this model, we simultaneously predict both axis-aligned and rotated-bounding boxes for the chutes, enabling us to detect the chute area in pouring videos.
The detected chute is represented by a bounding box with parameters $(x,y,w,h,\theta)$, indicating the center coordinates, width, height, and rotation angle of the box.

To enhance system efficiency, if the object detector continuously identifies the same region for more than $8$ frames, we halt the model's inference and fix the Region of Interest (RoI) in the video. Specifically, the RoI is determined by averaging the bounding box coordinates over the detected frames, ensuring stability in localization. This approach reduces GPU usage for every frame, thereby improving power efficiency while maintaining detection accuracy.

To focus on the chute interior, we convert the detected rotated bounding box into an upright rectangular patch before cropping. The resulting coordinates are then used to crop the unroated chute area, generating separate videos for the left section, $V^{L}$, and the right section, $V^{R}$, of the chute. 
By using these segmented videos, we extract only the information related to the concrete inside the chute, allowing the model to better understand the concrete features.

\subsection{Detecting Concrete Placement Location and Timing}
\label{subsec:step2}
On construction sites, concrete is discharged from two separate chutes connected to individual concrete mixer trucks. In this study, we detect which chute the concrete is falling from and accurately identify the starting point of the flow. This step is essential to ensure that the subsequent video classification model receives only the segments where concrete is actually flowing, enabling more precise slump analysis (see Step 2 in Figure~\ref{fig:pipeline}). Since video segments without concrete flow are not relevant for analysis, filtering out such irrelevant data in advance is a critical preprocessing step.

To precisely determine when and from which chute the concrete starts to fall, we perform motion analysis based on the segmented chute videos $V^L$ and $V^R$ obtained in Section~\ref{subsec:step1}. For each video segment, we extract bounding boxes for the chutes from the object detector and compute optical flow at the center point of each bounding box. Specifically, we apply the Lucas-Kanade method, a sparse optical flow algorithm, to track the motion vector $(u, v)$ at the center $(x, y)$ across consecutive frames. 
When the tracked center point crosses the bottom boundary of the bounding box over time, we identify the moment of concrete drop as well as the specific chute through which the concrete is discharged.

To determine whether the center point $(x_t, y_t)$ passes through the bottom edge of a rotated bounding box at the time $t$, we proceed as follows:

\subsubsection{Definition of the Bottom Edge Direction}
Given a rotated bounding box coordinates, the bottom edge lies perpendicular to the vertical axis of the box. The direction vector of the bottom edge is defined as:
\begin{equation}
    \mathbf{d}_{bottom} = 
    \begin{bmatrix}
        \cos (\hat{\theta}) \\
        \sin (\hat{\theta})
    \end{bmatrix}
    \quad \text{where, } \hat{\theta} = \frac{\theta \pi}{180}.
\end{equation}
Note that $\theta$ is converted from degrees to radians, $\hat{\theta}$, since trigonometric functions are defined in radians.
Next, we compute two endpoints $\mathbf{p}_1$ and $\mathbf{p_2}$ of the bottom edge by shifting from the center $(x,y)$ along the direction of the horizontal axis (which is orthogonal to the vertical axis) by half of the width:
\begin{equation}
    \mathbf{p}_{1} = 
    \begin{bmatrix}
        x \\ y
    \end{bmatrix}
     + \frac{w}{2}
     \begin{bmatrix}
         \cos (\hat{\theta}) \\ 
         \sin (\hat{\theta})
     \end{bmatrix},
     \quad
    \mathbf{p}_{2} = 
    \begin{bmatrix}
        x \\ y
    \end{bmatrix}
     - \frac{w}{2}
     \begin{bmatrix}
         \cos (\hat{\theta}) \\ 
         \sin (\hat{\theta})
     \end{bmatrix}          
\end{equation}

\subsubsection{Determining the Location and Timing}
To make the temporal change explicit, we define the center point of the tracked chute in each frame as $(x_{t}, y_{t})$ at time $t$, and as $(x_{t-1}, y_{t-1})$ at the previous frame $t-1$. The optical flow vector $(u,v)$ represents the displacement from frame $t-1$ to $t$, so that $(x_{t}, y_{t}) = (x_{t-1} + u, y_{t-1} + v)$. The endpoints of the bottom edge of the rotated bounding box, $\mathbf{p_1} = (x_{1}+\alpha, y_{1}+\alpha)$ and $\mathbf{p_2} = (x_{2}+\alpha, y_{2}+\alpha)$, are computed.
This translation mitigates false triggering caused by optical-flow sensitivity to shadows by slightly raising the decision boundary for a more stable crossing detection. We empirically use $\alpha=100$, applied as $+\alpha$ for the right chute and $-\alpha$ for the left chute.

To determine whether the tracked center point crosses the bottom edge between frames $t-1$ and $t$, we explicitly check the positions at both time steps relative to the edge.
First, the line equation for the bottom edge is:
\begin{equation}
    m = \frac{y_2 - y_1}{x_2 - x_1 + \epsilon}, \quad b = y_1 - mx_1,
\end{equation}
where $\epsilon$ is a small constant to avoid division by zero.
The signed vertical distance from the center point to the bottom edge at each time step is given by:
\begin{equation}
    d_{t} = y_{t} - (mx_{t} + b), \quad d_{t-1} = y_{t-1} - (mx_{t-1} + b).
\end{equation}
The center point is considered to have crossed the bottom edge if the signs of $d_{t}$ and $d_{t-1}$ are difference or zero:
\begin{equation}
    d_{t} \times d_{t-1} \leq 0.
\end{equation}
This approach allows us to identify both the active discharge chute and the timing of the concrete placement.

\subsection{Slump Prediction and Anomaly Detection}
To enable efficient and real-time estimation of concrete slump, we leverage the drop timing and location identified in Section~\ref{subsec:step2}. Specifically, given the drop time $t$, we predict the slump using video data from the corresponding chute region. This process corresponds to Step 3 of the overall pipeline shown in Figure~\ref{fig:pipeline}.
To focus on the continuous flow of concrete, we adopt a video classification model, ResNet 3D~\cite{tran2018:closer}. Let $V$ denote the chute video (i.e., $V^R$ or $V^L$), and $t$ the detected drop time.

\subsubsection{Data Preprocessing}
We crop the unrotated-bounding box corresponding to $V$ from the full image to isolate the chute area. From the frame at time $t$, we extract a sequence of $N$ consecutive frames from the cropped region to construct a temporally continuous input.

\subsubsection{Data Augmentation}
Given that construction sites are exposed to a wide range of lighting and environmental conditions, data augmentation is essential to improve the model’s generalization performance. To simulate these real-world variations, we employ a set of augmentation techniques, including ColorJitter, gamma correction, horizontal flipping, and contrast adjustment. We also apply MixUp~\cite{zhang2017:mixup} and CutMix~\cite{yun2019:cutmix}, which blend either the pixel content or the labels of two training examples to generate mixed samples. While MixUp interpolates both images and labels, CutMix replaces a region of an image with a patch from another image, adjusting the labels proportionally to the area. These strategies encourage the model to learn more robust and smoother decision boundaries, and to be less sensitive to spurious correlations between local image regions and class labels.

Together, these augmentations help the model handle diverse visual appearances, such as changes in brightness, shadows, occlusion, and orientation, that are commonly encountered in construction site environments. They also promote better regularization by expanding the training distribution and reducing overfitting, ultimately leading to improved performance under unseen conditions.

\subsubsection{Model}
We employed the ResNet 3D architecture~\cite{tran2018:closer} as our video classification backbone, designed to jointly capture spatial and temporal patterns. We experimented with two variants: Mixed Convolution and (2+1)D convolution.
The Mixed Convolution model uses 3D convolutions in early layers for motion modeling and 2D convolutions in later layers for efficient spatial reasoning, achieving performance comparable to full 3D models with fewer parameters.
The (2+1)D model factorizes 3D convolutions into separate spatial and temporal steps, enhancing optimization and model capacity.

\subsubsection{Training}
To formulate the slump estimation task as a classification problem, we categorized the continuous slump values into five categorical intervals (e.g. 150--180mm, 180--210mm), where the width of each interval reflects the $\pm30$ mm tolerance specified in the KS F 4009 standard~\cite{ksf4009}, along with an additional margin for human error.
For intervals at the extremes (i.e. below 150 mm and above 240 mm), which are rarely encountered in practice, we did not apply explicit error margins.

To enhance generalization, we applied the MixUp technique, which linearly interpolates both input videos and their corresponding labels to generate augmented samples with soft targets. Under this setting, we replaced the standard cross-entropy loss with a soft-target formulation that incorporates label smoothing:
\begin{equation}
    L_{cls} = -\lambda_{cls} \sum_{i=1}^{C} y_i^{soft} \log(\hat{y}_i),
\end{equation}
where $y_i^{soft} \in [0,1]$ denotes the soft label vector obtained via MixUp and label smoothing, and $\hat{y}_{i}$ represents the predicted probability for class $i$.
We empirically set $\lambda_{cls} = 0.1$.

By leveraging softened target distributions and encouraging smoother decision boundaries, this loss formulation helps mitigate overfitting to minority classes and reduces bias toward dominant categories—issues that frequently arise in imbalanced datasets. Consequently, the model is guided to generalize more effectively across all classes, regardless of their frequency in the training distribution.

\subsubsection{Slump Prediction and Anomaly Detection}
We use the trained video classification model to predict the slump of concrete.
To enhance both real-time prediction and robustness, we perform the prediction $T$ times using $N$ consecutive frames each time, and determine the final result via majority voting.
This allows for a more stable and reliable estimation of the slump.
Based on the predicted slump category, we compare it with the ordered slump range at the construction site. If the prediction falls within the specified range, the concrete is considered acceptable; otherwise, it is flagged as abnormal, enabling effective quality control of the material.

\section{Evaluation of Prediction Models}
\label{sec:experiments}
We perform both quantitative and qualitative evaluations at each step of the system to thoroughly assess the effectiveness of each component.
Specifically, we analyze how each module contributes to overall performance through ablation studies and visualization-based inspections.
\begin{figure*}[t]
    \centering
    \includegraphics[width=\textwidth]{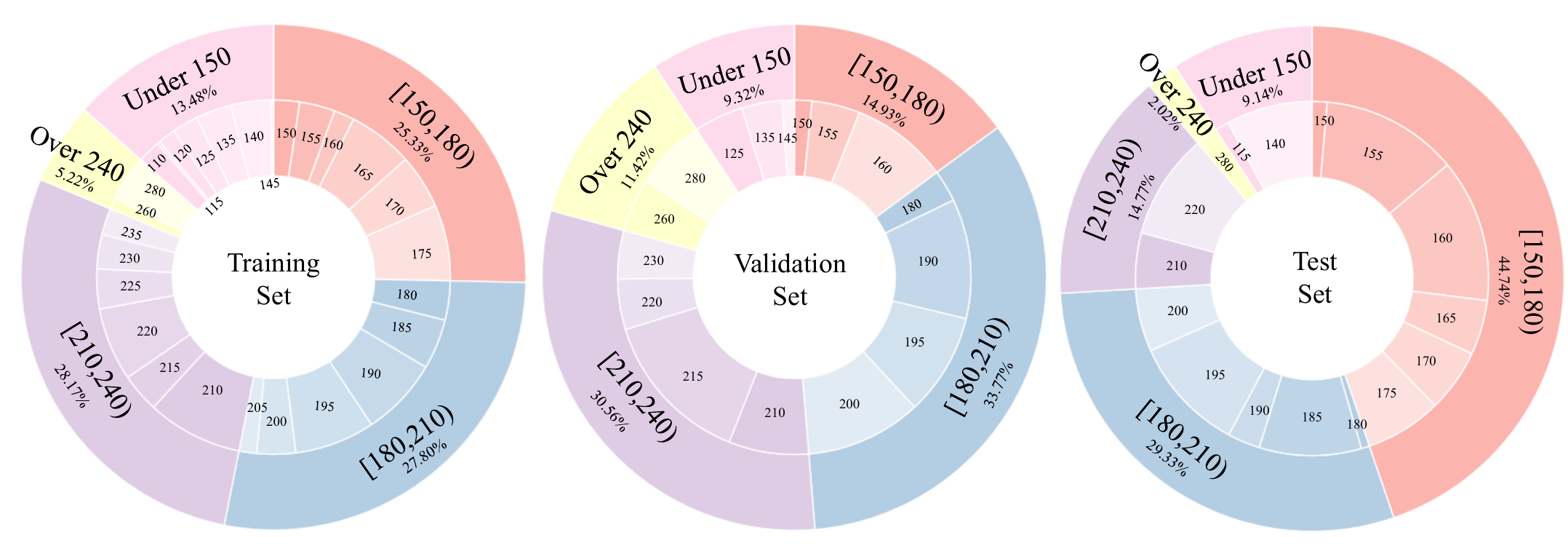}
    \caption{Statistics of our dataset. The outer pie shows the distribution of slump ranges, while the inner pie details the specific slump values within each category.}
    \label{fig:data_stat}
\end{figure*}

\subsection{Dataset}
Over an eight-month period, a dataset was constructed by replicating real-world construction site conditions in a controlled experimental settings. The recorded videos were segmented into uniform 10-second intervals, discarding any residual clips shorter than 10 seconds. This process resulted in $4,504$ training samples, $998$ validation samples, and $941$ test samples, as shown in Figure~\ref{fig:data_stat}.

\subsection{Chute Detection}
\subsubsection{Evaluation Metrics}
To evaluate the performance of chute detection, we adopt evaluation metrics commonly used in object detection tasks, namely mAP${}_{50-95}$ and precision.
Specifically, the mean average precision (mAP${}_{50-95}$) is computed as the average of AP values at IoU thresholds ranging from 0.50 to 0.95 in increments of 0.05:
\begin{equation}
    \text{mAP}_{50-95} = \frac{1}{10}\sum^{0.95}_{\text{IoU}=0.50}\text{AP}_{\text{IoU}}
\end{equation}
This metric provides a comprehensive measure of the detection accuracy at varying levels of localization strictness. Additionally, we report Precision, which quantifies the proportion of correctly predicted positive instances among all predicted positives:
\begin{equation}
    \text{Precision}=\frac{\text{TP}}{\text{TP} + \text{FP}}
\end{equation}
where TP and FP denote true positives and false positives, respectively. Together, these metrics offer a robust assessment of both the detection quality and reliability of the chute localization module.

\subsubsection{Results}
\begin{figure*}[t]
    \centering
    \includegraphics[width=\textwidth]{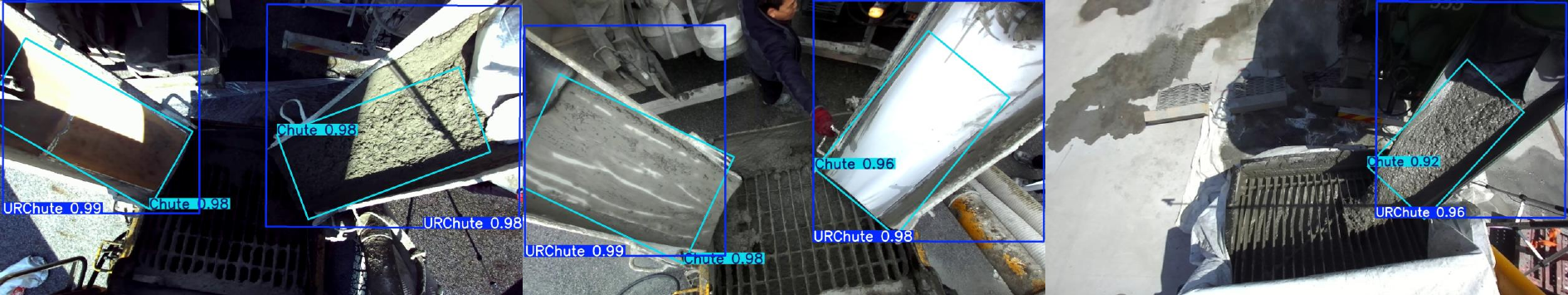}
    \caption{Qualitative results of chute detection. Blue and Sky blue bounding boxes represent detections of ``URChute'' and ``Chute'', respectively. The number inside each box denotes the confidence score.}
    \label{fig:yolo_qual}
\end{figure*}

Our chute detection model achieved strong quantitative results, with an mAP${}_{50-95}$ of \textbf{0.9945} and a precision of \textbf{0.995}. The qualitative results are shown in Figure~\ref{fig:yolo_qual}, where all bounding boxes exhibit confidence scores above $0.95$, indicating highly reliable detections.
This demonstrates the strong and effective performance of our chute detection model.
Furthermore, real-time inference is feasible, as the model achieves an average speed of $9.0$ms per frame, making it suitable for real-time applications.

\subsection{Detecting Concrete Placement Location}
\begin{table}[t]
\centering
\resizebox{0.75\textwidth}{!}{%
\begin{tabular}{@{}c|ccccc:c@{}}
\toprule
                           & Under 150 & 150 $\sim$180 & 180 $\sim$210 & 210 $\sim$240 & Over 240 & Avg   \\ \midrule
\multicolumn{1}{c|}{Left}  & 100       & 100           & 100           & 100           & 98.45      & 99.91 \\
\multicolumn{1}{c|}{Right} & 98.60     & 95.00         & 91.90         & 88.70         & 95.19      & 92.89 \\
\multicolumn{1}{c|}{None}  & 100       & 100           & 100           & 100           & 100        & 100   \\ \hdashline
\multicolumn{1}{c|}{Avg}  & 99.30     & 97.68         & 95.40         & 94.51         & 96.92      &       \\ \bottomrule
\end{tabular}%
}
\caption{Results of concrete placement location across various slump conditions and chute position. ``None'' means no pouring situation. All values in the table are in accuracy (\%).}
\label{tab:chute_location}
\end{table}
To evaluate which chute of the mixer truck the concrete was poured from, human annotators manually labeled the ground truth for each instance.
Identifying the chute from which the concrete is poured also allows for accurate determination of the pouring start time, thereby enabling more precise quantitative evaluation.
To facilitate this evaluation, we conducted experiments under three scenarios: pouring from the left-side chute, pouring from the right-side chute, and no pouring.
This result is shown in Table~\ref{tab:chute_location}.

As shown in the Table~\ref{tab:chute_location}, the proposed system demonstrated an overall accuracy exceeding $90\%$ on average, and achieved $100\%$ accuracy in cases where concrete placement did not occur.
While the chute located on the left side exhibited $100\%$ accuracy in most experimental cases, several errors were observed for the chute on the right side. This was primarily attributed to the shadow cast by the hopper cover located above the right-side chute.
In particular, it was observed that after approximately 10 seconds from the start of the video, the direction of the optical flow vectors $(u,v)$ tended to follow the movement of the shadow rather than the actual flow of the concrete. However, such errors are not expected to pose significant issues in practical field applications when determining whether concrete is being placed.
Additionally, in samples with slump values exceeding 240 mm, a temporary decline in accuracy was also observed for the left-side chute. This phenomenon occurred when the slump was excessively high, resulting in smooth and fluid concrete flow, which made pixel-level motion detection using optical flow more difficult. This issue was observed only in the sample with a slump value of 280 mm.

These experimental results demonstrate that the proposed system can accurately detect active chutes with high reliability. Since chute identification directly corresponds to detecting the start of concrete placement, the system is proven to be effective for automatic recognition of the placement zone and estimation of the starting time in real-world construction environments.

\subsection{Slump prediction}
\subsubsection{Model Training Setup}
We trained the model using 16-frame video clips, sampled at a frame interval of 2, and extended by a temporal factor of 2. The resize ratio was randomly selected from \{ 3/4, 4/3 \}, and standard data augmentations were applied. For spatial augmentation, we used random horizontal flipping with a probability of 0.5 and ColorJitter with the following ranges: brightness 0.4, contrast 0.4, saturation 0.4, and hue 0.1. Additionally, MixUp ($\alpha$ = 0.2, probability = 0.5) and CutMix ($\alpha$ = 0.2, probability = 0.5) were employed to generate soft-label training samples, encouraging better generalization and robustness to noisy data. Label smoothing with a factor of 0.1 was also applied.

The model was optimized using the AdamW optimizer~\cite{loshchilov2017:decoupled} with a learning rate of $1e-4$, and a weight decay of $1e-4$. We used a total batch size of 128 and trained for 10 epochs using the OneCycle learning rate scheduler~\cite{smith2019:super} with cosine annealing, with a peak learning rate set to $1e-3$.

\begin{table}[t]
\centering
\begin{minipage}[t]{0.49\textwidth}
\centering
\resizebox{\textwidth}{!}{%
\begin{tabular}{@{}|cc|cc|cc@{}}
\toprule
                                                                                    &              & \multicolumn{2}{c|}{Val} & \multicolumn{2}{c}{Test} \\ 
Backbone                                                                            & Architecture & Acc         & F1         & Acc         & F1         \\ \midrule
\multicolumn{1}{c|}{Transformer}                                                   & TimeSFormer~\cite{bertasius2021:space}  & 0.6980      & 0.7280     & 0.5724      & 0.6155     \\ \hdashline
\multirow{3}{*}{ResNet 3D} & MC${}_3$-18     & 0.8080      & 0.8069     & 0.7916      & 0.8526     \\
\multicolumn{1}{c|}{}      & R3D-18          & 0.8152      & \textbf{0.8183}     & 0.8011      & 0.8511     \\
\multicolumn{1}{c|}{}      & R(2+1)D-18      & \textbf{0.8152}      & 0.8149     & \textbf{0.8226}      & \textbf{0.8691}     \\ \bottomrule
\end{tabular}%
}
\caption{Evaluation results across various video classification model architectures.}
\label{tab:architecture}
\end{minipage}
\hfill
\begin{minipage}[t]{0.49\textwidth}
\centering
\resizebox{\textwidth}{!}{%
\begin{tabular}{@{}ccc|cc@{}}
\toprule
Label smoothing & Weighted sample & MixUp & Acc    & F1     \\ \midrule
\checkmark      &                &                 & 0.7654 & 0.8180 \\
\checkmark      &               & \checkmark     & 0.7717 & 0.8205 \\
\checkmark      & \checkmark      &              & 0.7955 & 0.8499 \\
\checkmark      & \checkmark      & \checkmark     & \textbf{0.8226} & \textbf{0.8691} \\
\bottomrule
\end{tabular}%
}
\caption{Ablation study results on the training strategy for R(2+1)D-18. ``\checkmark'' indicates that the technique was applied, while a blank cell denotes that it was not.}
\label{tab:ablation}
\end{minipage}
\end{table}

\subsubsection{Results} We evaluate our model’s ability to predict the range of concrete slump using both the validation and test sets. Accuracy and F1-score are used as the primary evaluation metrics. In addition to evaluating the overall performance, we conduct an ablation study to assess the effectiveness and robustness of each component of our proposed model.

As shown in Table~\ref{tab:architecture}, our models employing ResNet-3D as the backbone achieve over 80\% accuracy on both the validation and test sets. In contrast, the Transformer~\cite{vaswani2017:attention}-based TimeSFormer~\cite{bertasius2021:space} exhibits lower accuracy.
Among the evaluated architectures, we selected R(2+1)D-18 as our final model, as it achieved the highest performance on both validation and test sets. Based on this architecture, we conducted a detailed ablation study, and the corresponding results are reported in Table~\ref{tab:ablation}.

To mitigate data imbalance and prevent overfitting, we incorporated label smoothing, weighted sampling, and MixUp-based data augmentation. Each of these techniques led to measurable performance gains, as demonstrated in our experiments.
Through these experiments, we successfully developed an optimized and robust model for concrete slump prediction.

\begin{figure*}
    \centering
    \includegraphics[width=\linewidth]{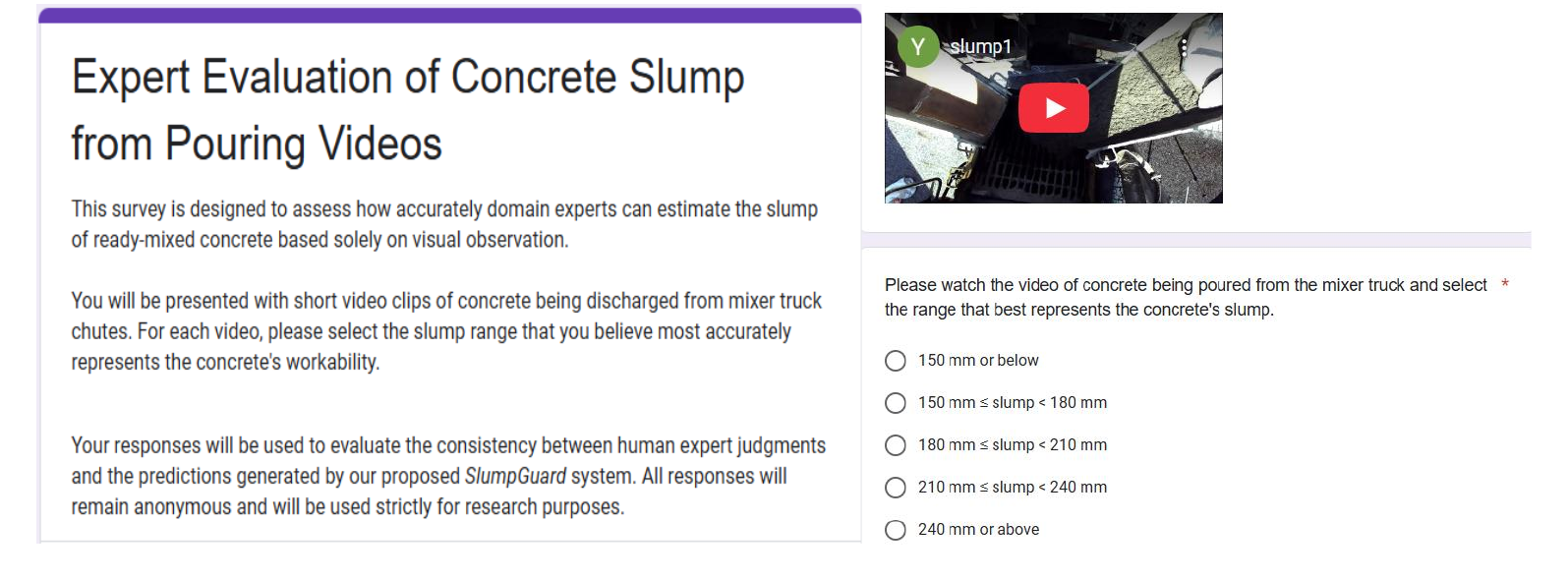}
    \caption{Screenshot of the expert evaluation interface used in our study. A Google Forms–based questionnaire was provided to domain experts, requiring them to choose a single slump interval from five predefined categories. }
    \label{fig:humaneval}
\end{figure*}
\subsubsection{Expert Evaluation}
To further validate the practical reliability of \textsc{SlumpGuard} under real-world operational conditions, we conducted an expert human evaluation to assess how accurately people can visually estimate concrete slump.
For this evaluation, we asked 11 domain experts with over 15 years of field experience to review ten video samples of concrete discharging from mixer trucks and estimate the corresponding slump range based solely on visual observation.
All selected videos were cases for which \textsc{SlumpGuard} produced correct predictions, and each clip shown to the experts consisted of a 10-second segment.
A screenshot of the Google Forms interface used for this expert evaluation is shown in Figure~\ref{fig:humaneval}, with the results summarized in Tables~\ref{tab:expert_acc} and Figure~\ref{fig:confusion}.

Despite being highly experienced professionals, the experts exhibited notable disagreement when visually estimating slump ranges. As reported in Table~\ref{tab:expert_acc}, only 9.1–-45.5\% of the experts correctly identified the ground-truth slump in each sample, and no video exhibited unanimous agreement among all evaluators. Moreover, the confusion matrix in Figure~\ref{fig:confusion} indicates that most misclassifications occurred between adjacent slump ranges (e.g., 180–210 mm vs. 210–240 mm), revealing the inherent difficulty of visually distinguishing small differences in concrete workability.
These findings highlight the subjective variability and inconsistency of human judgment in real-site conditions, even among experts with extensive field experience. In contrast, \textsc{SlumpGuard} provided correct predictions for all evaluated samples, demonstrating its potential as a reliable and objective tool for real-time slump estimation without the need for manual testing.

\begin{table}[t]
\centering
\begin{tabular}{c c}
\begin{minipage}{0.48\linewidth}
\centering
\caption{Expert evaluation for slump estimation. Ground-truth slump values and the number (percentage) of correct responses among eleven experts.}
\begin{tabular}{c c c}
\toprule
\textbf{GT (mm)} & \textbf{Correct (11)} \\
\midrule
Under 150 (150)        & 4 (36.4\%) \\
180 $\sim$ 210 (200)       & 3 (27.3\%) \\
Under 150 (145)      & 5 (45.5\%) \\
180 $\sim$ 210 (190)      & 5 (45.5\%) \\
Over 240 (280)         & 5 (45.5\%) \\
210 $\sim$ 240 (280)      & 1 (9.1\%)  \\
150 $\sim$ 180 (170)   & 3 (27.3\%) \\
180 $\sim$ 210 (190)  & 3 (27.3\%) \\
150 $\sim$ 180  (165) & 4 (36.4\%) \\
210 $\sim$ 240  (220)   & 3 (27.3\%) \\
\bottomrule
\label{tab:expert_acc}
\end{tabular}
\end{minipage}
&
\begin{minipage}{0.52\linewidth}
\centering
\includegraphics[width=\linewidth]{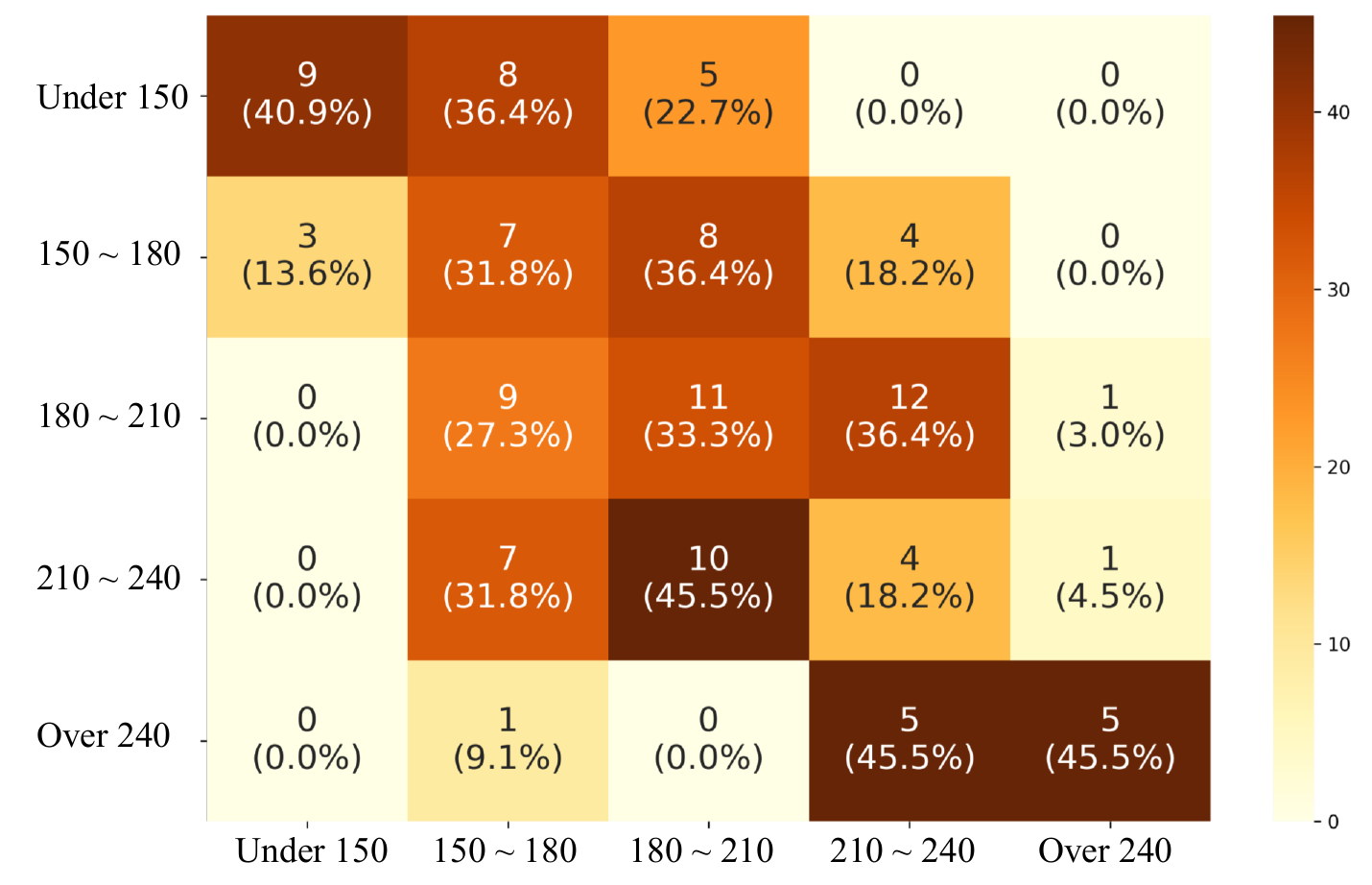}\\
\captionof{figure}{Confusion matrix of expert predictions. Numbers represent counts and percentages relative to the ground-truth slump class.}
\label{fig:confusion}
\end{minipage}
\end{tabular}
\end{table}

\section{Real-World Deployment}
\label{sec:deployment}
\begin{figure}
    \centering
    \includegraphics[width=\linewidth]{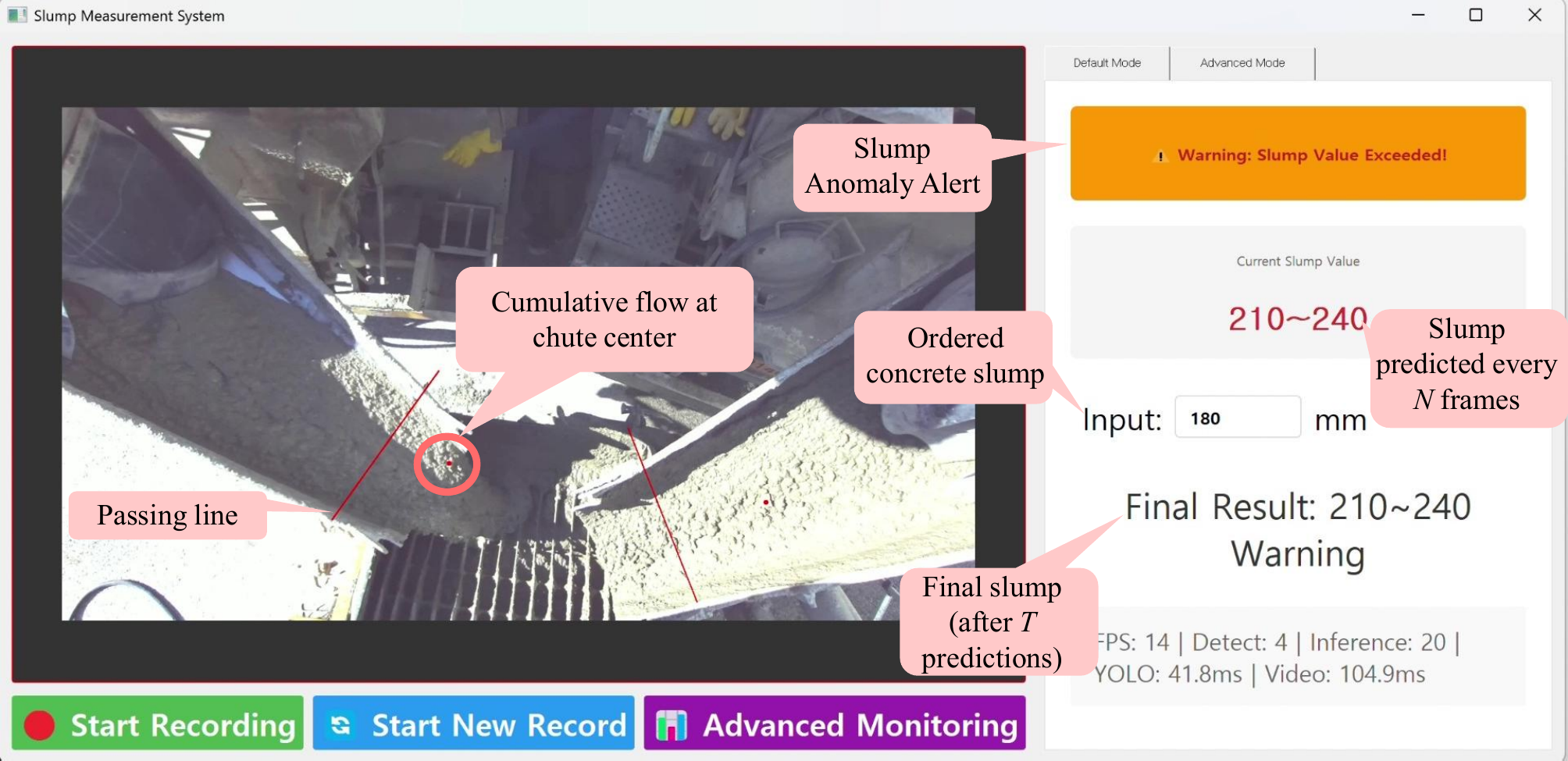}
    \caption{Screenshot of the deployed \textsc{SlumpGuard} user interface. The left panel visualizes chute detection, the passing-line based flow gating, and cumulative flow at the chute center. The right panel displays the system’s slump prediction (updated every $N$ frames), the ordered slump value, and the final majority-voted slump result. A warning message is triggered when the predicted range does not match the ordered specification.}
    \label{fig:ui}
\end{figure}
To demonstrate the practical applicability of \textsc{SlumpGuard} in real construction environments, we present the on-site operational user interface (UI) used during deployment, as shown in Figure~\ref{fig:ui}.

Before concrete placement begins, the ordered slump specification for each batch is entered into the system (see ``Input'' in Figure~\ref{fig:ui}).
When two mixer trucks arrive at the placement area, the system automatically detects both chutes without any manual intervention. It then identifies which of the two trucks is actively discharging concrete and determines the precise onset of pouring (see red circle and passing line in Figure~\ref{fig:ui}). Once the pouring onset is detected, the slump classification model is activated and begins predicting the slump category by every $N$ frames (see ``Current Slump Value'' in Figure~\ref{fig:ui}). The system produces predictions over ten successive inference cycles (i.e., $T=10$), after which the final slump value is determined through majority voting. Based on this prediction, the system compares the inferred slump range with the ordered specification for the corresponding batch. If the prediction deviates from the specified range, the UI displays a warning flag; otherwise, the batch is marked as acceptable.

This interface is lightweight, requires no specialized hardware beyond a fixed camera, and integrates seamlessly into existing construction workflows.
This deployed interface illustrates how \textsc{SlumpGuard} functions as a fully autonomous, practical tool for continuous and non-intrusive quality assurance on site, offering an interpretable and actionable output that supports real-time decision-making during concrete placement.




\section{Discussion}
\label{sec:discussion}
This section discusses four key aspects of the proposed system: (1) the architectural design of the modular pipeline, (2) the system-level implications for real-world deployment, (3) the practical constraints imposed by actual construction workflows, and (4) the expert-level evaluation that contextualizes the model’s performance relative to human judgment.

\subsection{Pipeline Design}
The three-stage structure adopted in this work provides several advantages for field deployment. Separating the tasks of chute localization, pouring-event detection, and slump classification allows each component to operate under well-defined assumptions, improving interpretability and enabling targeted refinement when system behavior must be analyzed or adapted to a new site. The motion-gated mechanism ensures that only slump-relevant temporal windows are processed, reducing computational cost and supporting real-time operation. Although an end-to-end formulation may appear conceptually simpler, it is structurally unsuitable for this task. An end-to-end model would need to implicitly learn the pouring onset, accurately localize the chute under large geometric and rotational variation, and simultaneously infer motion cues required for flow-based gating. These sub-tasks demand fundamentally different spatial and temporal inductive biases, which cannot be reliably integrated into a single unified network. By contrast, the modular pipeline offers greater flexibility, maintainability, and transparency, which are essential properties for real-world operation in dynamic on-site environments.

\subsection{System-level Implication}
The purpose of \textsc{SlumpGuard} extends beyond its modular pipeline design to its ability to function as a fully autonomous system that aligns with practical construction-site workflows. Sensor-based approaches often encounter operational challenges, such as restrictions on hardware installation, differences in responsibility between concrete suppliers and contractors, and additional labor requirements for dedicated monitoring.
In contrast, the proposed system requires only a single fixed camera and does not depend on any mixer-truck-mounted equipment, allowing it to be integrated seamlessly into standard concrete placement procedures. This vision-based, non-intrusive design enables deployment across a wide range of site conditions while offering scalability and practical utility for continuous quality monitoring.

Furthermore, explicit modeling of pouring onset and the interpretability of intermediate outputs facilitate inspection of system behavior under various environmental conditions and support the identification of potential error sources, thereby enhancing operational trust. These characteristics demonstrate that \textsc{SlumpGuard} is a field-ready solution that offers the reliability, transparency, and maintainability required for real-world construction environments. They also illustrate how an integrated vision-based pipeline can serve as a practical alternative or complement to conventional manual or sensor-based inspection processes.

\subsection{Practical Constraints in Real Construction Workflows}
The proposed system reflects the practical constraints under which slump inspection is conducted in real construction environments. In standard concrete placement, mixer trucks are required to remain stationary on level ground, and discharging from a tilted vehicle is not permitted due to safety regulations and the risk of altering fresh concrete properties. Therefore, ground-slope variation does not represent a realistic condition for slump evaluation in practice.

In addition, slump inspection is not carried out at night or under unstable illumination, as concrete placement and visual quality assessment are mandated to occur under sufficiently lit conditions. Since \textsc{SlumpGuard} operates within these same constraints and focuses analysis on the interior of the chute region, unrealistic scenarios such as steep slopes or nighttime pouring fall outside the practical scope of slump evaluation.

\subsection{Expert-level Evaluation}
The expert study provides additional insight into how visual slump assessment is performed in practice and helps contextualize the model’s performance relative to human judgment. Notably, the variability in expert responses indicates that visual estimation of slump is inherently subjective. In real construction workflows, slump is rarely determined by closely observing the concrete as it flows out of the mixer truck. Instead, field engineers typically inspect the concrete after it has accumulated on the placement surface, relying on experiential cues developed through repeated exposure to that specific condition. As a result, the experts in our study were asked to evaluate a visual scenario that differs from their usual inspection context, namely, assessing slump directly from the discharge stream rather than from the already deposited material.

This mismatch between familiar field practice and the evaluation setting likely contributed to the lower-than-expected agreement among experts. Because their judgments depend heavily on tacit knowledge and accumulated impression-based cues, the absence of familiar reference conditions can lead to divergence in interpretation. The expert-level evaluation thus reflects the inherent ambiguity associated with purely visual slump estimation and underscores the practical challenge of achieving consistent human judgment in such tasks.

Taken together, these findings suggest that the predictive consistency offered by \textsc{SlumpGuard} may provide meaningful support in situations where visual assessment alone leads to substantial inter-operator variability. This further highlights the potential role of automated, vision-based systems in complementing traditional quality assurance workflows.

\section{Limitations}
\label{sec:limitation}
This study has several limitations. First, the optical-flow–based motion detection is sensitive to shadows and illumination changes, particularly within darker or low-contrast regions of the chute interior, where flow vectors may become distorted. However, this effect is largely limited to the frame selection and pouring-event detection stages rather than directly impacting the accuracy of slump classification. Furthermore, the use of data augmentation techniques during training contributes to improving the illumination robustness of the final classification model.
Second, although the dataset was collected in a controlled experimental environment designed to closely replicate real concrete pouring operations, it does not originate from fully uncontrolled and actively operating construction sites. Additional validation across a broader spectrum of site conditions, including varying equipment standards, operational workflows, and environmental interference, will be necessary to confirm broader applicability.
Third, the system assumes a fixed viewpoint in which the interior region of the chute remains clearly visible. If the camera position is unstable, if the chute is occluded by workers, or if the capturing angle significantly deviates from standard placement, the performance of detection and flow computation may degrade. Finally, the dataset primarily reflects construction practices and equipment configurations common in Korea. Further domain adaptation and validation are required to ensure generalizability to regions with different mixer truck designs, operational standards, or construction procedures.

\section{Conclusion}
\label{sec:conclusion}
This paper presented \textsc{SlumpGuard}, an automated system for monitoring and classifying concrete slump using video analysis. 
The proposed approach employs a three-stage pipeline, consisting of chute detection, pouring location and timing estimation, and concrete slump classification. 
Specifically, a YOLO-based object detection model was developed to identify the concrete mixer truck and its discharge chute. Based on these detections, optical flow analysis was employed to accurately determine the pouring location and timing. Finally, a video classification model was used to predict the concrete slump, enabling automated and comprehensive inspection of all incoming concrete deliveries, thereby replacing manual sampling procedures and contributing to more reliable quality assurance in concrete construction. Conclusions are drawn as follows.
\begin{enumerate}
    \item To facilitate video-based concrete slump prediction, we constructed a dataset comprising 6,443 video clips, each 10 seconds long, capturing various concrete pouring scenes. Bounding box annotations for the discharge chute of the mixer truck were provided to enable the development of deep learning-based methods.
    \item The proposed YOLO-based chute detector demonstrated highly reliable performance in identifying the discharge chute of the mixer truck. It achieved a mean Average Precision (mAP${}_{50:95}$) of 99.45 and a Precision of 99.5, indicating a high level of accuracy.
    \item We developed an algorithm to determine the location and timing of concrete discharge using the computer vision technique of optical flow. Experimental results demonstrated an average accuracy of over 95\%.
    \item We showed that the concrete slump range can be automatically predicted using a video classification model. Through a series of comparative experiments and ablation studies, we identified the optimal model architecture and hyper-parameters, achieving a high-accuracy slump prediction model with a performance of 82\%.
    Based on these results, our system can automatically verify whether the predicted slump range matches the ordered specification, enabling the development of an automated quality control system for concrete.
\end{enumerate}

Future work could focus on predicting continuous concrete slump values rather than categorical ranges. In addition, based on the results of this study, the system could be integrated with Building Information Modeling (BIM) or construction process management systems to enable automated tracking of concrete quality throughout the entire construction lifecycle, ultimately leading to a more practical and deployable solution.

\section*{Acknowledgements}
This work was supported by the Institute of Information \& Communications Technology Planning \& Evaluation (IITP) grant funded by the Korea Government (MSIT) (No. RS-2024-00354218, No. RS-2020-II201361, Artificial Intelligence Graduate School Program at Yonsei University), and by the Ministry of Land, Infrastructure and Transport of the Korea Government through the Railway Technology Research Program (No. RS-2021-KA163289).
\section*{Declaration of competing interest}
The authors declare that they have no known competing financial interests or personal relationships that could have appeared to influence the work reported in this paper.
\appendix




\bibliographystyle{elsarticle-num} 
\bibliography{ref}




\end{document}